\documentclass{article}

\usepackage{arxiv}

\usepackage{amsmath}		
\usepackage[utf8]{inputenc} 
\usepackage[T1]{fontenc}    
\usepackage{hyperref}       
\usepackage{url}            
\usepackage{booktabs}       
\usepackage{amsfonts}       
\usepackage{nicefrac}       
\usepackage{microtype}      
\usepackage{cleveref}       
\usepackage{lipsum}         
\usepackage{graphicx}
\usepackage{natbib}
\usepackage{doi}

\usepackage{comment}
\usepackage{float}
\usepackage{gensymb}
\usepackage{multirow}
\usepackage{xcolor}
\usepackage{url}

\title{Evaluation of Dirichlet Process Gaussian Mixtures for Segmentation on Noisy Hyperspectral Images}

\author{
	\href{https://orcid.org/0000-0002-3377-7591}
	{\includegraphics[scale=0.06]{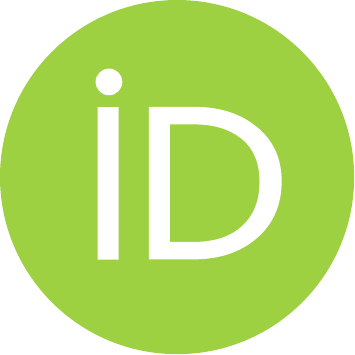}\hspace{1mm}
	Kiran Mantripragada} \thanks{corresponding author.} \\
	Visual Computing Lab - Faculty of Science\\
	Ontario Tech University\\
	2000 Simcoe Street North, Oshawa, ON L1G OC5, Canada \\
	\texttt{kiran.mantripragada@ontariotechu.net} \\
	\And
	\href{https://orcid.org/0000-0002-8992-3607}{\includegraphics[scale=0.06]{figs/orcid.pdf}\hspace{1mm}Faisal Z. Qureshi}\\
	Visual Computing Lab - Faculty of Science\\
	Ontario Tech University\\
	2000 Simcoe Street North, Oshawa, ON L1G OC5, Canada \\
	\texttt{faisal.qureshi@ontariotechu.ca} \\
}

\renewcommand{\shorttitle}{HSI segmentation with DPGMM}

\DeclareRobustCommand\onedot{\futurelet\@let@token\@onedot}
\def\@onedot{\ifx\@let@token.\else.\null\fi\xspace}
 
\def\ie{\emph{i.e}\onedot}

\def\etal{\emph{et al}\onedot}

\DeclareMathOperator*{\argmax}{arg\,max}

\hypersetup{
pdftitle={Evaluation of Dirichlet Process Gaussian Mixtures for Segmentation on Noisy Hyperspectral Images},
pdfsubject={cs.CV, cs.ML},
pdfauthor={Kiran Mantripragada},
pdfkeywords={Hyperspectral, segmentation, DPGMM, Dirichlet Process, Gaussian Mixture Model},
}

\begin{document}
\maketitle

\begin{abstract}
  Image segmentation is a fundamental step for the interpretation of Remote Sensing Images. Clustering or segmentation methods usually precede the classification task and are used as support tools for manual labeling. The most common algorithms, such as k-means, mean-shift, and MRS, require an extra manual
  step to find the scale parameter. The segmentation results are severely affected if the parameters are not
  correctly tuned and diverge from the optimal values. Additionally, the search for the optimal scale is a
  costly task, as it requires a comprehensive hyper-parameter search. This paper proposes and
  evaluates a method for segmentation of Hyperspectral Images using the Dirichlet Process Gaussian Mixture
  Model. Our model can self-regulate the parameters until it finds the optimal values of scale
  and the number of clusters in a given dataset. The results demonstrate the potential of our method to find
  objects in a Hyperspectral Image while bypassing the burden of manual search of the optimal parameters. In addition, our model also produces similar results on noisy datasets, while previous research usually
  required a pre-processing task for noise reduction and spectral smoothing.
\end{abstract}


\section{Introduction}
\label{introduction}
Image segmentation is an essential step before the primary tasks such as object detection and classification.
Haralick and Shapiro mentioned in their seminal paper \cite{haralick:1985} that the clustering process can be
viewed as segmentation. Several authors use these terminologies interchangeably, but it is also usual to differ
on how the grouping method is done: segmentation is done on the spatial domain of the image, while clustering
is done on the measurement space. Additionally, ``semantic segmentation'' is a classification process in
which we assign object identifications to each pixel.

In the context of Hyperspectral Images (HSI) processing, spatial features such as borders and textures
become less relevant due to the richness of spectral features. Analysis on HSI usually seeks for one of two
tasks: object detection and materials identification~\cite{borzov:2018,signoroni:2019,heylen:2014}. Semantic segmentation, which explores spatial
features, is more prevalent for object detection, while ``pixel-unmixing'', which uses spectral features, is
a required step prior to the identification of materials. To keep the terminology concise, we will
use the terms ``Semantic  Segmentation'' for object detection and ``Clustering'' for arbitrary pixel groupings. Such terminology is also used previously as in~\cite{garcia:2022,hamida:2017,kemker:2018}.

\begin{figure}
    \includegraphics[width=0.355\linewidth, trim=0cm 0cm 0cm 0cm, clip=true]{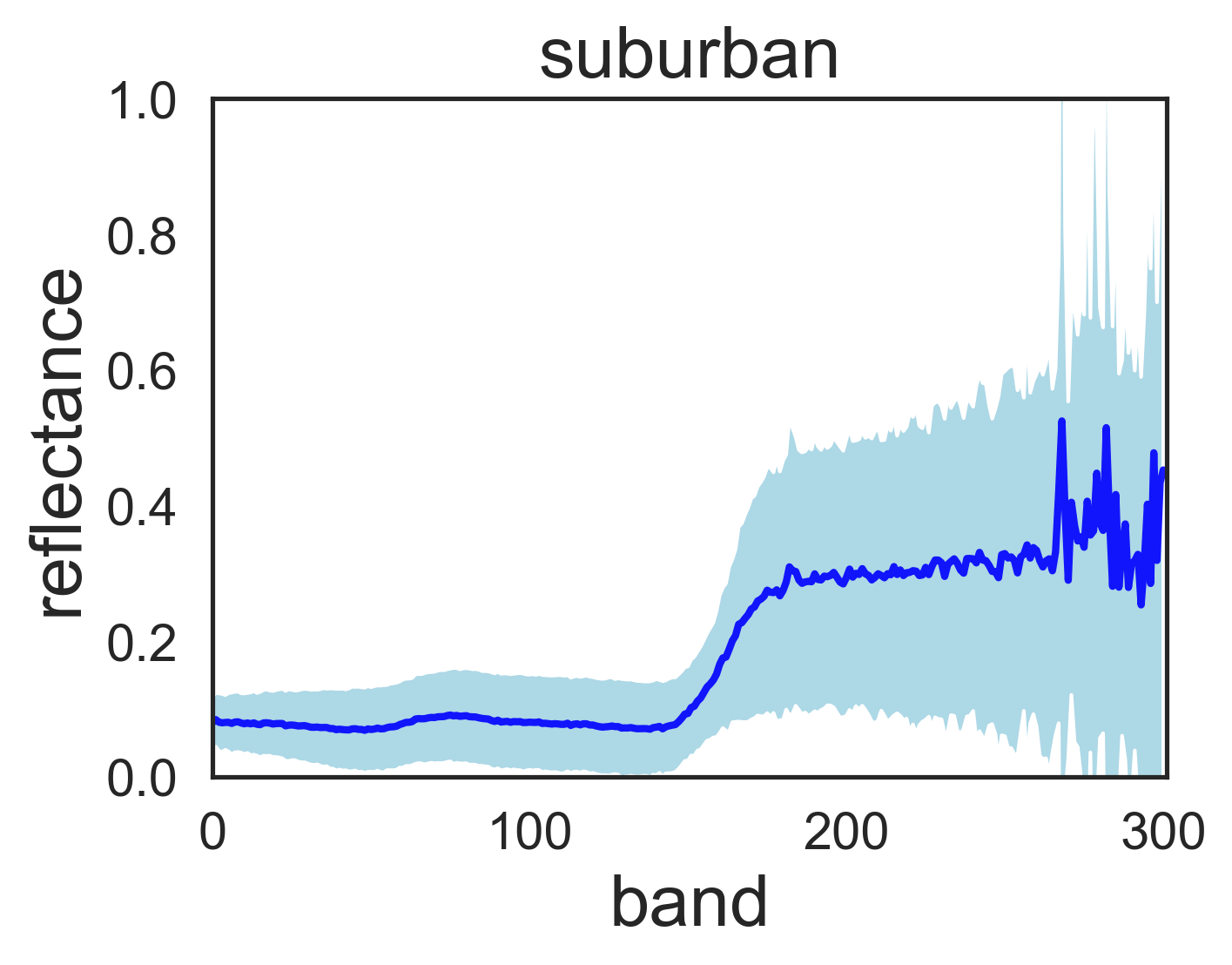}
    \includegraphics[width=0.30\linewidth, trim=1.9cm 0cm 0cm 0cm, clip=true]{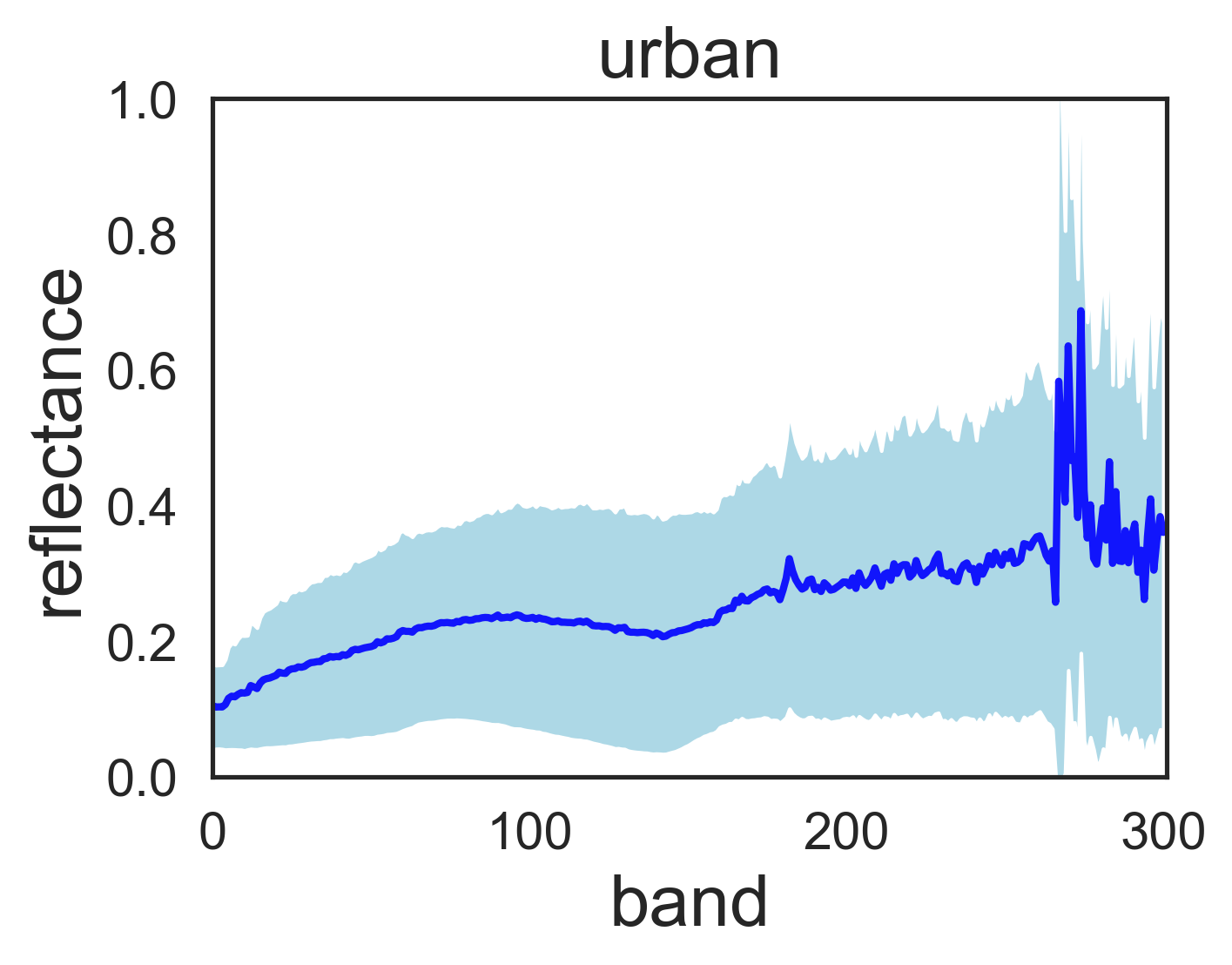}
    \includegraphics[width=0.30\linewidth, trim=1.9cm 0cm 0cm 0cm, clip=true]{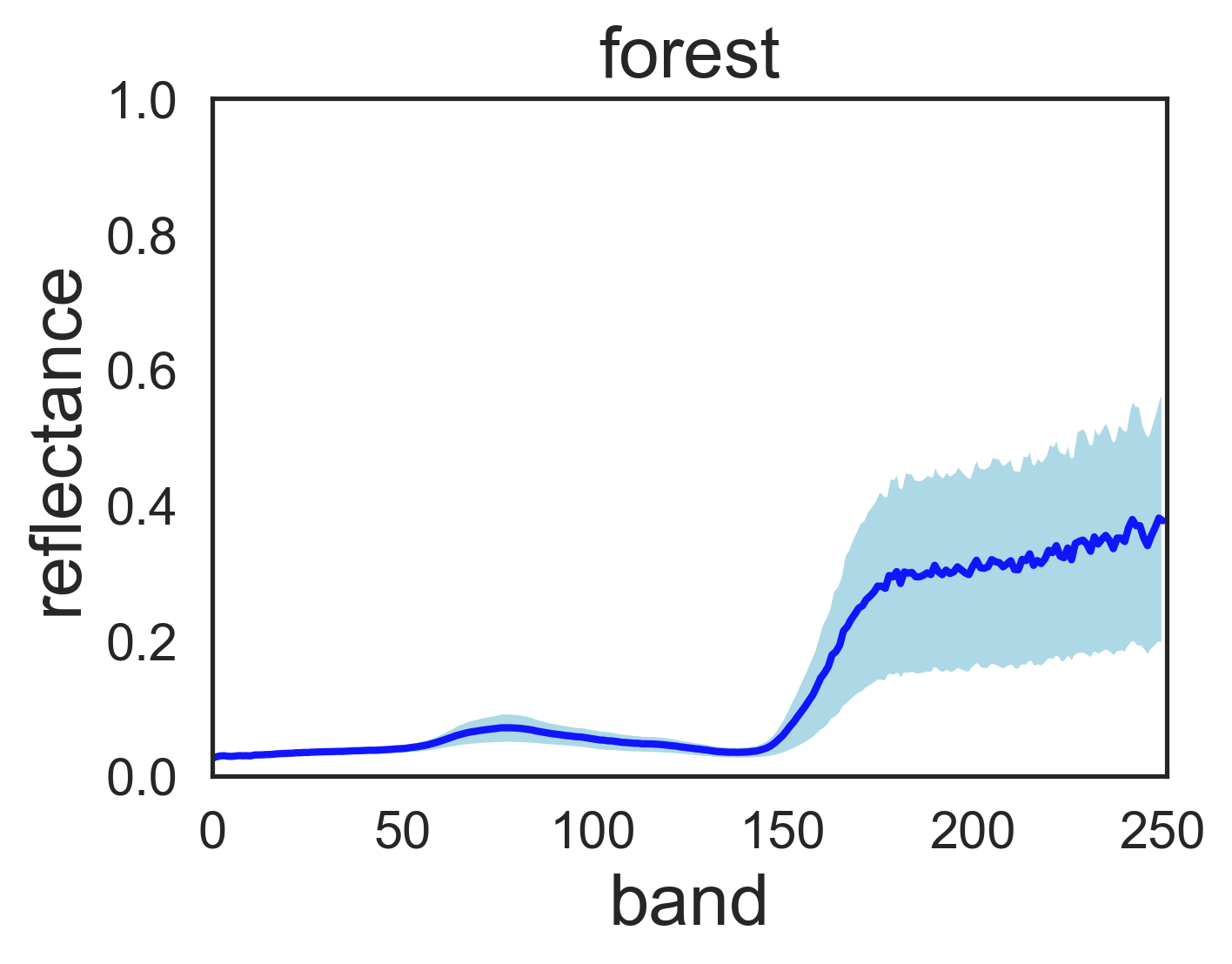}
    \caption{Average of all pixels for each dataset. Higher bands are more susceptible to noise which impacts negatively the segmentation results. From Left  right: Suburban, Urban, and Forest.}
    \label{fig:average_pixels}
  \end{figure}

In this paper, we propose a method for HSI clustering using Dirichlet Process Gaussian Mixture Model (DPGMM).
We choose this model for two main reasons: 1) investigate the separability of a hyperspectral pixel into independent signals\footnote{analogous to Blind Source Separation (BSS) methods}
and their respective proportions in the mix; and 2) define a model that can capture the structure
and variability of HSI pixels while also being robust to the noise present in remote sensing data.
Figure~\ref{fig:average_pixels} shows the average spectral curve with the variability and
noise per channel.

It has been demonstrated that GMMs can generate complex multimodal distributions that successfully capture
the structure of data \cite{gorur:2010,wu:2016,nascimento:2012}, and such versatility of GMMs have been
explored before in earth observation, remote sensing, and hyperspectral image processing
~\cite{wu:2016,prasad:2014,nascimento:2012,zare:2008}. However, to the extent of our knowledge, the research
community lacks a quantitative comparison of DPGMM against the previous and most prestigious segmentations
algorithms. Also, different from existing methods, we propose a solution based on Variational Inference in a
neural network setting to learn the parameters of the DPGMM.

We compute the DPGMM model by ``minimizing the divergence of conjugate priors and posteriors'' given the
data and we use the Variational Inference (VI) method to solve the optimization problem. In order to evaluate
the performance of DPGMM for clustering, we compare against existing segmentation and clustering methods
recently published in the literature~\cite{dao:2021}. The contribution of this research are three-fold: 1) we
develop and evaluate a clustering method for HSI datasets using DPGMM; 2) we use three datasets previously
explored in ~\cite{dao:2021} (see Section~\ref{sec:datasets}) and make them publicly available; and 3) we
open-source our implementation of DPGMM for easier reproducibility and further exploration by the HSI and
remote sensing community.

\section{Related work}
\label{related_work}

\subsection{Clustering methods for RGB images}
Clustering of pixels in RGB images has been a very active research topic since the early days of computer vision
until today. Relevant work dates back from the early 80's, such as the seminal publication of Haralick and Shapiro \cite{haralick:1985}, until today, with lots of impressive results from
Deep Learning research, such as \cite{minaee:2021,caron:2020,hossain:2019,kemker:2018,hamida:2017,wang:2018}.

Clustering of regular (3-channel RGB) images has been extensively studied in computer vision. Broadly speaking, the existing methods can be divided into three categories: (1) edge-based methods; (2)
region-based methods; and (3) pixel-descriptors methods
~\cite{hossain:2019,dao:2021,zhou:2019,yin:2015,long:2015,wang:2005,zhang:2013}. Overall, the methods (1)
and (2) are directly applicable to RGB or grayscale images, although they require adaptation to work with
hyperspectral images (HSI). Most of the approaches used in RGB images leverage spatial contextual
information because distinctive descriptors of pixels cannot rely only on the spectra that contain mostly
color information. Therefore the extraction of spatial features such as borders and texture is crucial for
RGB image classification tasks. Publications in remote sensing RGB image processing such as
\cite{garcia:2022, hamida:2017, kampffmeyer:2016, kemker:2018, wang:2018} explore Deep Learning, while
some authors explore methods such as k-means~\cite{dao:2021,zhai:2021,caron:2018},
watershed~\cite{dao:2021,pooja:2015}, SLIC (Simple Linear Iterative Clustering)~\cite{achanta:2010}, and
mean-shift~\cite{dao:2021,greggio:2012}.

\subsection{Clustering methods for HSI}
Clustering for HSI can be implemented using a variety of methods for pixel descriptors. One can incorporate
spatial, spectral, or the conjunction of spatial+spectral features. Several authors, such as
\cite{al-khafaji:2022, feng:2022, li:2021, zhang:2021c}, explore the use of spectral and spatial features
combined. Despite sometimes being beneficial, the richness of features can also bring the higher
computational burden, higher processing times, the Hughes phenomenon, while supervised methods also require
large amounts of ground-truth data for training. Combining these issues can make the solution
not viable in some applications; therefore, the usability of all the features starts to be questionable. To address these problems, some authors also explore dimensionality reduction methods in HSI
clustering and classification~\cite{ahmad:2019, cahill:2014, datta:2018}.

For scenarios where the main task is ``material identification'' through pixel-unmixing, we
interpret that spatial context is less relevant because a single pixel can represent several materials. Therefore the transition (borders) between objects is not evident or detectable by border-detector filters. In this context, we say that each pixel is composed of a mixture of pure
endmembers\footnote{Endmember is the spectral signature of a single material}.
Conversely, in very high-resolution images, every single pixel represent a unique material that might be
more close to the pure endmember, therefore providing more cues about the independent signal of each
material.
Because the Remote Sensing-based HSI is collected at high altitudes, the image resolution is naturally
low, \ie each pixel represent a combination of one or more materials.

\subsection{Probability and Mixture Models-based clustering}
  Mixture Models have been extensively used in image classification and segmentation.
  \cite{greggio:2012, nguyen:2013, yu:2010} use GMM (Gaussian Mixture Models) for segmentation on RGB images.
  \cite{shah:2004} explored the Independent Component Analysis Mixture Model (ICAMM) to solve the problem of separability of endmembers. The model is implemented using the mutual information-maximization learning
  algorithm. However, it is assumed the mixture is a linear combination of the components.
  Similar to our work, \cite{acito:2003} propose a segmentation algorithm using GMM. However, it is not clear
  how the model is solved or trained. The number of channels is drastically reduced by simply removing the
  channels with higher noise. Nascimento \etal propose a GMM model to solve the problem of unmixing under the
  same assumptions of our work: Dirichlet distributions automatically enforce the sum-to-one and
  non-negativity constraints. The model is computed thorough the iterative expectation-maximization algorithm.
  The main differences in our work rely on the fact that we solve the DPGMM model through variational inference
  in a neural network setting.
\section{Method}
\label{sec:method}
\subsection{Dirichlet Process Gaussian Mixture Model}

\newcommand{\sig}{\ensuremath{\boldsymbol{\Sigma}}}
\newcommand{\muj}{\ensuremath{\boldsymbol{\mu}_j}}
\newcommand{\sigj}{\ensuremath{\boldsymbol{\Sigma}_j}}
\newcommand{\im}{\ensuremath{\boldsymbol{X}}}
\newcommand{\pix}{\ensuremath{\boldsymbol{x}}}

Consider a hyperspectral image $\im$ with $N$ pixels: $\pix_1,\pix_2,\cdots,\pix_N$, where $\pix_i \in \mathbb{R}^D$.  We seek to represent this image as Gaussian Mixture Model (GMM) with $K$ components.  Under this regime likelihood of $\pix_i$ is
\begin{equation}
    \mathcal{L}(\Theta |\pix_i) = \sum_{j=1}^{K}\pi_j \; \operatorname{Normal}(\pix_i | \muj, \sigj),
\end{equation}
where $\operatorname{Normal}$ is the Multivariate Normal Distribution, $\muj$ and $\sigj$ are the mean and variance  for the $j^\mathrm{th}$ component, respectively, and $\pi_j$ is the fractional contribution of each component.  The vector of fractional contributions $\boldsymbol{\pi}$ is a $K$-simplex vector, i.e., $\sum_j \pi_j = 1$ and $\pi_j \ge 0$.  $\muj \in \mathbb{R}^D$ and $\sigj \in \mathbb{R}^{D \times D}$.  $\Theta$ is a placeholder for $\pi_j$, $\muj$ and $\sigj$ for $j \in [1,K]$.

We seek to estimate $\Theta$ by minimizing the overall negative log-likelihood
\begin{equation}
    - \log \mathcal{L}(\Theta | X) = - \sum_{i=1}^N \log \mathcal{L}(\Theta | \pix_i).
    \label{nll}
\end{equation}
Note that parameters $\pi_j$ and $\sigj$ take special forms: values $\pi_j$ must meet sum-to-one and non-negativity constraints and $\sigj$ are covariance matrices and these must be symmetric and semi-positive definite.  This suggests that it is not sufficient simply minimize the loss in Eq.~\ref{nll}.  We also need to define auxiliary losses or regularizing terms to constrain these parameters appropriately.  We achieve this by defining priors for these parameters.   Samples drawn from a Dirichlet distribution satisfy the $K$-simplex nature of vector $\boldsymbol{\pi}= \{\pi_1,\pi_2,\cdots,\pi_K \}$.  Therefore, let
\begin{eqnarray*}
  \boldsymbol{\pi} & \sim & \operatorname{Dirichlet}\left(\dfrac{\{\alpha_1, \cdots,\alpha_K\}}{K}\right), \mathrm{\ where} \\
  \alpha_j & \sim & \operatorname{InverseGamma}(1,1).
\end{eqnarray*}
For the sake of computational efficiency, we assume that each (pixel) channel is independent and identically distributed.  Thus, $\sigj = \operatorname{diag}(\sigma_j^1,\sigma_j^2,\cdots,\sigma_j^D)$
 and an appropriate prior for $\sigj$ is the Inverse Gamma distribution.  Let
\begin{eqnarray*}
  \sigj & \sim &\operatorname{InverseGamma}(1,1).
\end{eqnarray*}
Additionally, we assume
\begin{eqnarray*}
  \boldsymbol{\muj} & \sim & \operatorname{Normal}(0,1).
\end{eqnarray*}
For more details on our choice of priors, please refer to work by  G{\"o}r{\"u}r  \etal~\cite{gorur:2010}, Deisenroth~\cite{deisenroth:2020}, Gelman~\cite{gelman:1995}, and Mathal~\cite{mathal:1992}).

It is possible to solve the minimization problem defined in Eq.~\ref{nll} within a variational inference setting, e.g., by defining a Kullback-Leibler divergence loss using the priors on $\pi_j$, $\muj$ and $\sigj$.  We also tried this approach first; however, we noticed slow and numerically unstable convergence behavior.  Instead we use the priors to construct negative log-likelihood values for
 $\pi_j$, $\muj$ and $\sigj$.  We found that this approach works better in practice.  Putting it all together, parameters $\Theta$ of the Gaussian Mixture Model are estimated by minimizing the following loss term
 \begin{align}
  l(X ; \Theta) = & - \log \mathcal{L}(\Theta | X) \\
                              & - \log p_{\pi}(\pi | .) \\
                              & - \log p_{\mu}(\mu | .) \\
                              & - \log p_{\sig}(\sig | .),
\end{align}
where $p_{\pi}$, $p_{\mu}$ and $p_{\sig}$ are priors defined above.

At inference time, $\pix_i$ is classified into one of $K$ clusters as follows
\begin{equation}
    c_i = \argmax_{j \in [1, K]} \operatorname{Normal}(\pix_i | \muj, \sigj),
\end{equation}
where $c_i$ denotes the cluster for pixel $\pix_i$.







\subsection{Segmentation Metrics}
Once we have a trained DPGMM model, we run the segmentation algorithm, \ie ~inference mode, of the DPGMM, on
the three datasets (Section ~\ref{sec:datasets}).
As also explored in the previous work of Dao \etal~\cite{dao:2021} where the authors compared the results
against a comprehensive set of scales and different algorithms, we evaluate our method using the commonly
used metrics $\operatorname{OS}$ (over-segmentation), $\operatorname{US}$ (under-segmentation), and
$\operatorname{ED}$ (Euclidean distance between  $\operatorname{OS}$ and  $\operatorname{US}$).

\begin{align}
  \operatorname{OS}_{i,j} & = 1- \frac{area(r_i \cap s_j)}{area(r_i)} \label{eq:os}                                       \\
  \operatorname{US}_{i,j} & = 1- \frac{area(r_i \cap s_j)}{area(s_j)} \label{eq:us}                                       \\
  \operatorname{ED}_{i,j} & = \sqrt{ \dfrac{\operatorname{US}_{i,j}^{2} + \operatorname{OS}_{i,j}^{2} }{2}} \label{eq:ed}
\end{align}
where $r_i \in R$ is the area (in pixels) of the ground-truth polygon $i$ (as depicted in Figure~\ref{fig:study_area}),  $s_j \in S$ is the area of segment $j$
computed by the algorithms.

However, due to the capability of DPGMM to automatically ``find'' the number of clusters, we compare only
with the optimal scale selected in the work of Dao \etal ~\cite{dao:2021}, in which, the scales were manually
selected using the ROC (Rate of change) curves of variance.
\section{Hyperspectral datasets}
\label{sec:datasets}


\begin{figure}
    \centerline{
        \includegraphics[width=.98\linewidth]{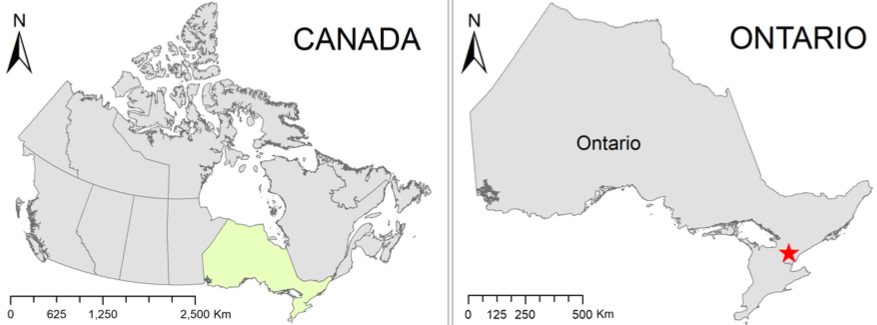}
    }
    \vspace{0.2cm}
    \centerline{
        \includegraphics[width=0.98\linewidth]{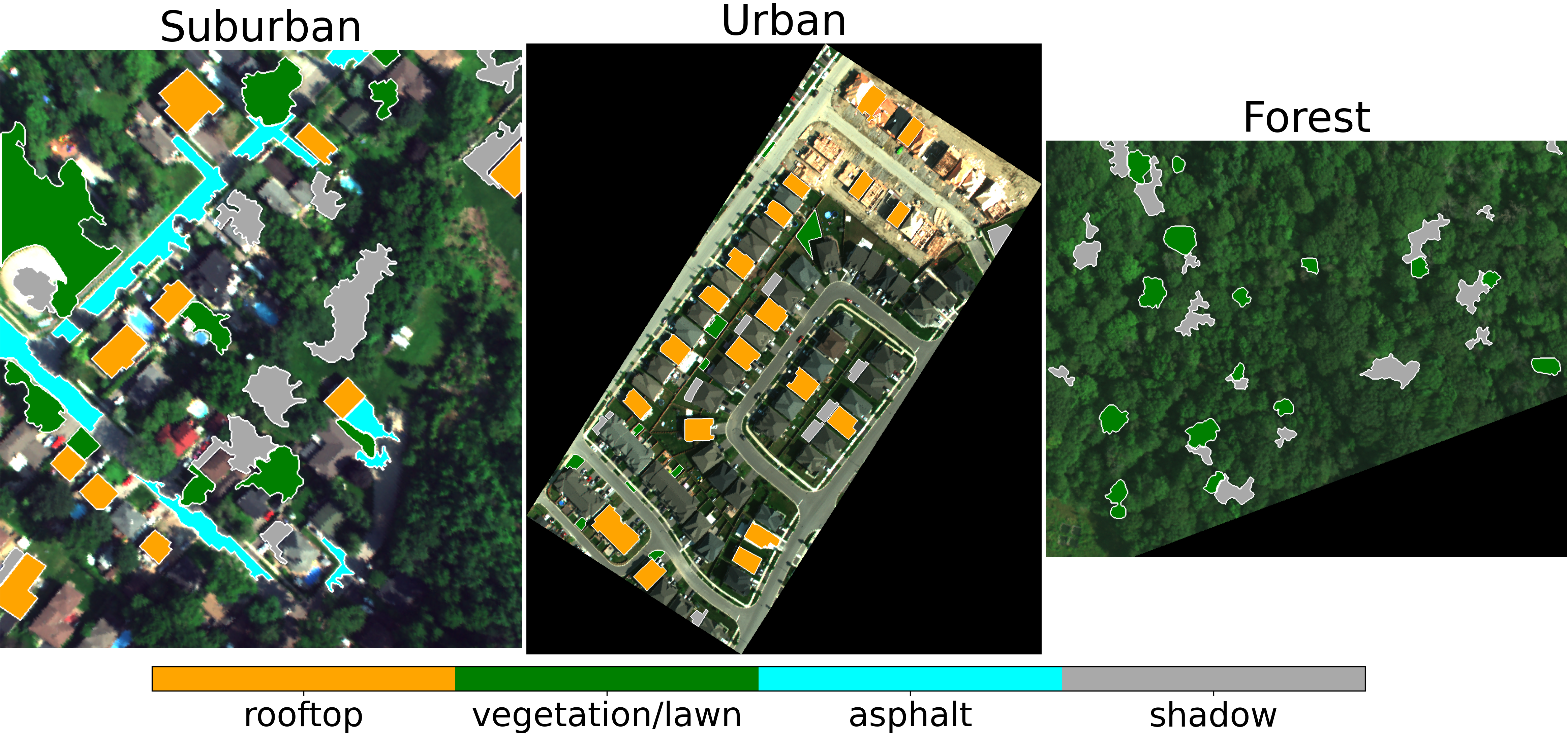}
    }
    \caption{The hyperspectral datasets were collected using an airborne
        sensor by the Remote Sensing and Spatial Ecosystem Modeling (RSSEM)
        laboratory, the Department of Geography, Geomatics and
        Environment, University of Toronto around the Toronto region
        (depicted by the red star) in Ontario, Canada. The bottom row
        shows the three datasets in pseudocolor (RGB images). This
        visualization was constructed using the 670 nm (red), 540 nm
        (green), and 470 nm (blue) bands from original data. The yellow,
        green, blue, and gray polygons overlaid on the hyperspectral images
        are the areas with ground-truth pixel labels available.}
    \label{fig:study_area}
\end{figure}

We used three high spatial resolution hyperspectral images for the studies presented in this paper (Figure~\ref{fig:study_area}). These images were captured using the Micro-HyperSpec III sensor (from Headwall Photonics Inc., USA) mounted at the bottom of a helicopter. The images were captured during the
daytime at 10:30 am on August 20, 2017. The original images with 325 bands were resampled to obtain 301 bands from 400 nm to 1000 nm with an interval of 2 nm. Raw images were converted to at-sensor radiance using
HyperSpec III software.

The images were also atmospherically corrected to surface reflectance using the empirical line calibration
method~\cite{dao:2019} with field spectral reflectance measured by FieldSpec 3 spectroradiometer from Malvern
Panalytical, Malvern, United Kingdom. These images represent 1) urban, 2) transitional suburban, and 3)
forests landcover types. These three landcover types cover a large fraction of use cases for hyperspectral
imagery; urban and sub-urban images are often used for city planning and land use analysis. Forest images are typically used for forest management, ecological monitoring, and vegetation analysis. The overlaid
polygons in Figure~\ref{fig:study_area} depict the annotated regions for which ground-truth
pixel labels are available. Figure~\ref{fig:study_area} (second row, left) shows the hyperspectral] image=
collected in an urban-rural transitional area. We refer to this image as the ``Suburban'' dataset. It was
captured around the Bolton area in southern Ontario and covers an area between $43\degree 52'32''$ and $43\degree
    53'04''$ in latitude and $-79\degree 44'15''$ and $-79\degree 43'34''$ in longitude. This region consists of
various land cover types, such as rooftops, asphalt roads, swimming pools, ponds, grassland, shrubs, and urban
forest. The image also contains regions that are in shadows. The image resolution is $0.3$ square meters
, and the covered area is around $41,182$ square meters.

Figure~\ref{fig:study_area} (second row, middle) shows the hyperspectral image collected in a residential
urban area, also around the Bolton region in southern Ontario. We refer to this image as the ``Urban'' dataset. It
contains rooftops, under-construction residences, roads, and lawns landcover types. The dataset also exhibits
regions that are in shadows. This image covers the area between $43\degree 45'30''$ and $43\degree 45'43''$
in latitude and $-79\degree 50'06''$ and $-79\degree 49'51''$ in longitude. The image resolution is $0.3$
square meters, and the area after removing background pixels is around $59,834$ square meters.

Figure~\ref{fig:study_area} (second row, right) shows the hyperspectral dataset collected in a natural forest
located at a biological site of the University of Toronto in the King City region in southern Ontario. We refer
to this dataset as the ``Forest'' dataset. It covers the area between $ 44\degree 01' 58'' $ and $44\degree
    02'04''$ in latitude and $-79\degree 32'06''$ and $-79\degree 31'55''$ in longitude.  The image resolution is
$0.3$ square meters, and the area after removing background pixels is around $43,084$ square meters.
\section{Experiments and Results}
\label{results}

The mathematical abstraction of DPGMM allows for an infinite number of classes. However, we limited the
maximum number of clusters $max(K)=5$ in our experimental set because the number of categories present in our
datasets is at most 4. Being an unsupervised model, we trained the model using the entire datasets to learn
the parameters of the distribution $\mathcal{L}(\boldsymbol{x};\boldsymbol{\pi}, \boldsymbol{\mu},
\boldsymbol{\sigma})$.  Once trained, the model estimates the likelihood of a new pixel belonging to any of the clusters.

We used the $\operatorname{OS}$, $\operatorname{US}$ and $\operatorname{ED}$ (Equations
\ref{eq:os}, \ref{eq:us}, and \ref{eq:ed} respectively) for a quantitative measurement of
the segmentation. The smaller the values, the better the quality of segmentation.

The existing labeled samples used to measure the segmentation results provide one class per pixel instead of a mix of materials. Therefore we can fairly compare with previous segmentation methods that look
assigns a single class per pixel.

\begin{figure}[!t]
  \centering
  \includegraphics[width=0.49\linewidth]{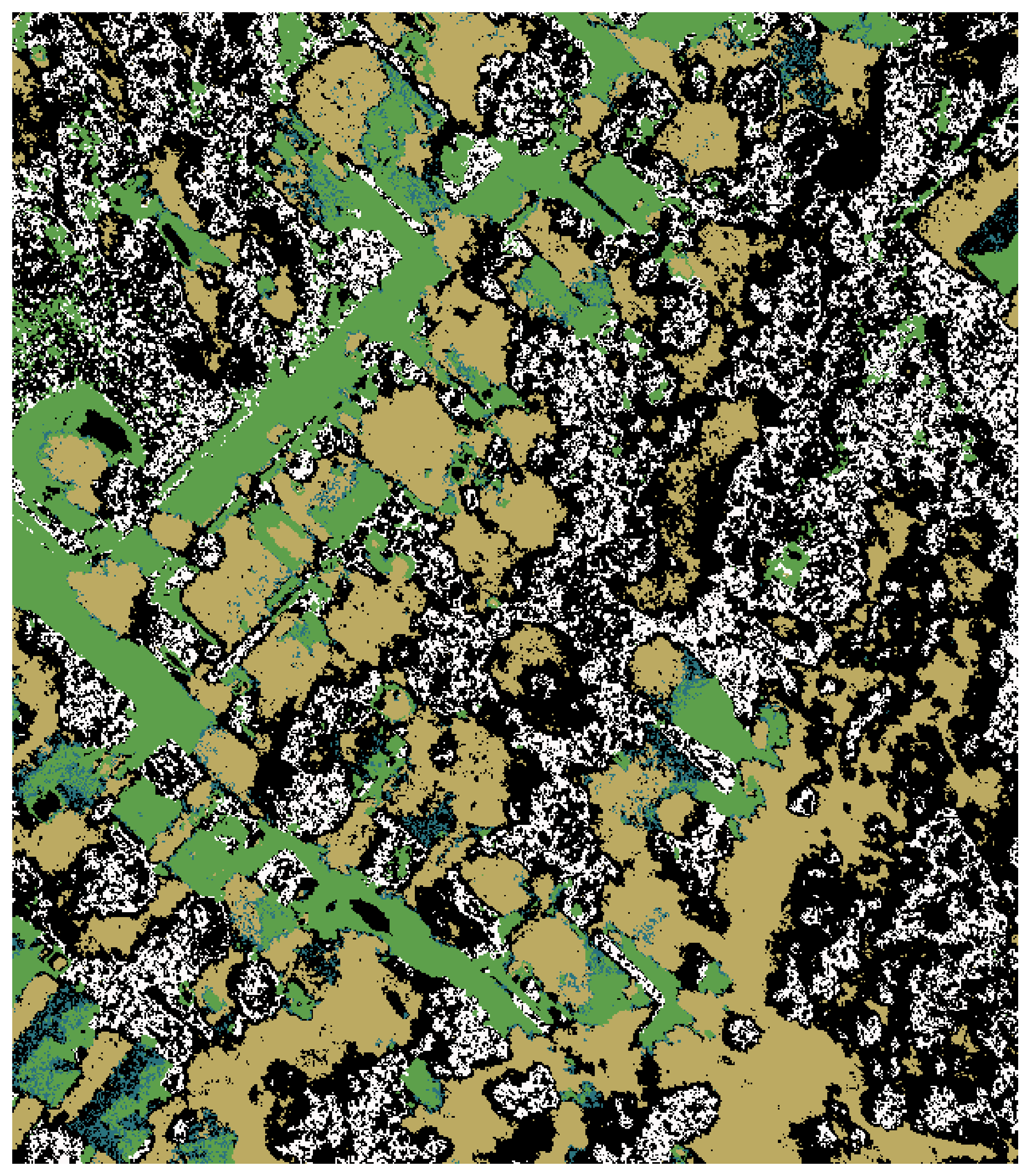}
  \includegraphics[width=0.49\linewidth]{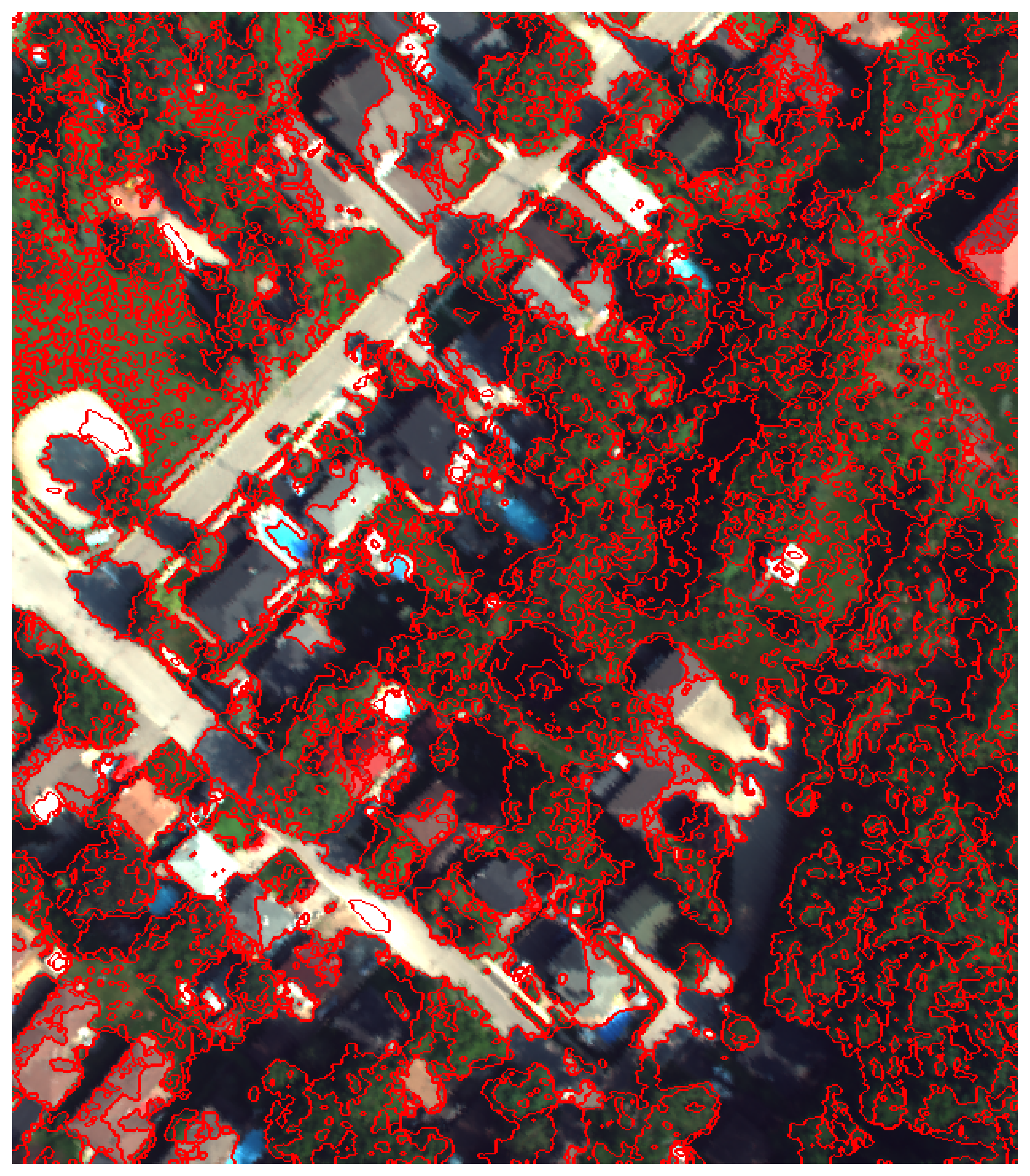}
  \includegraphics[width=0.49\linewidth]{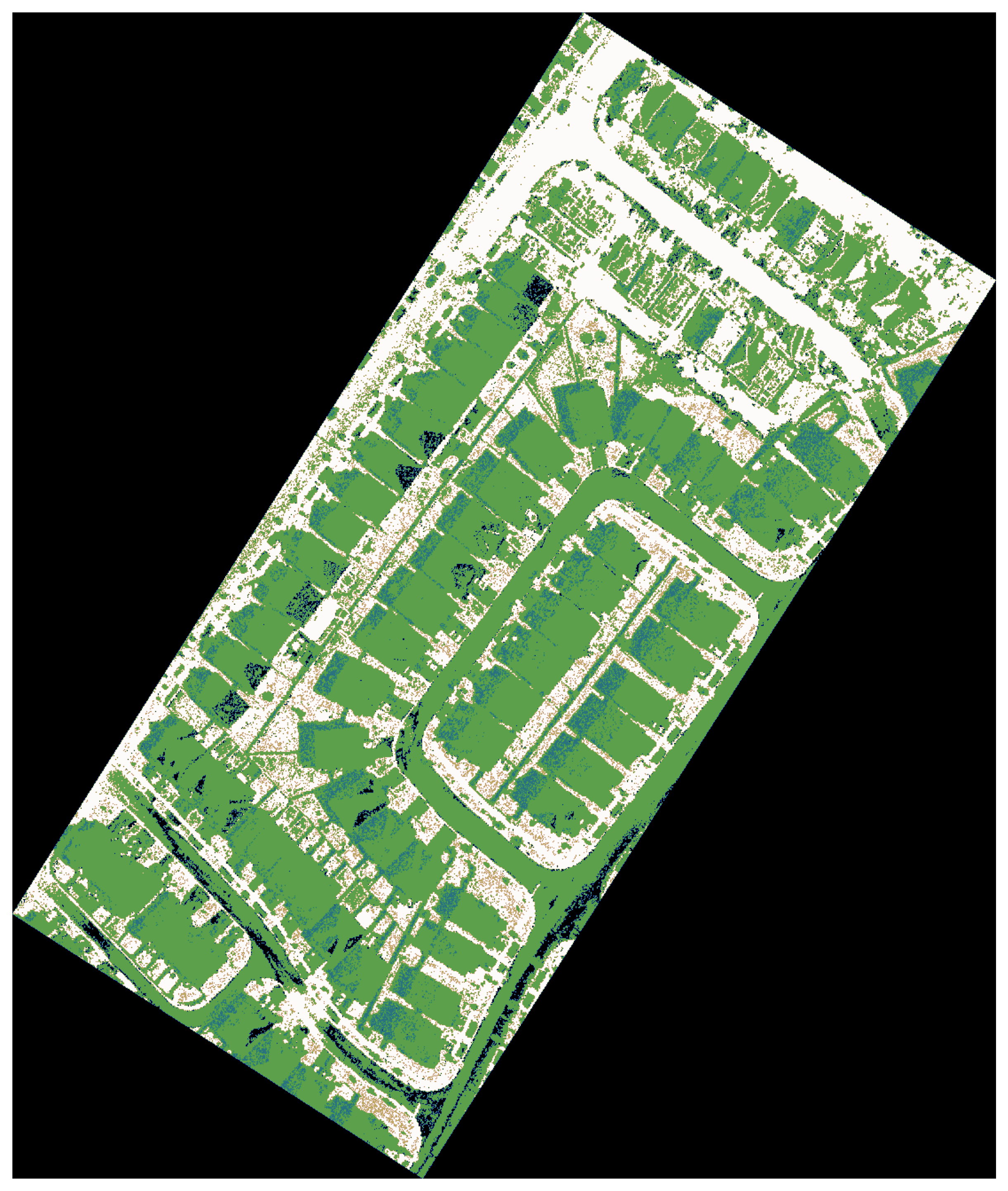}
  \includegraphics[width=0.49\linewidth]{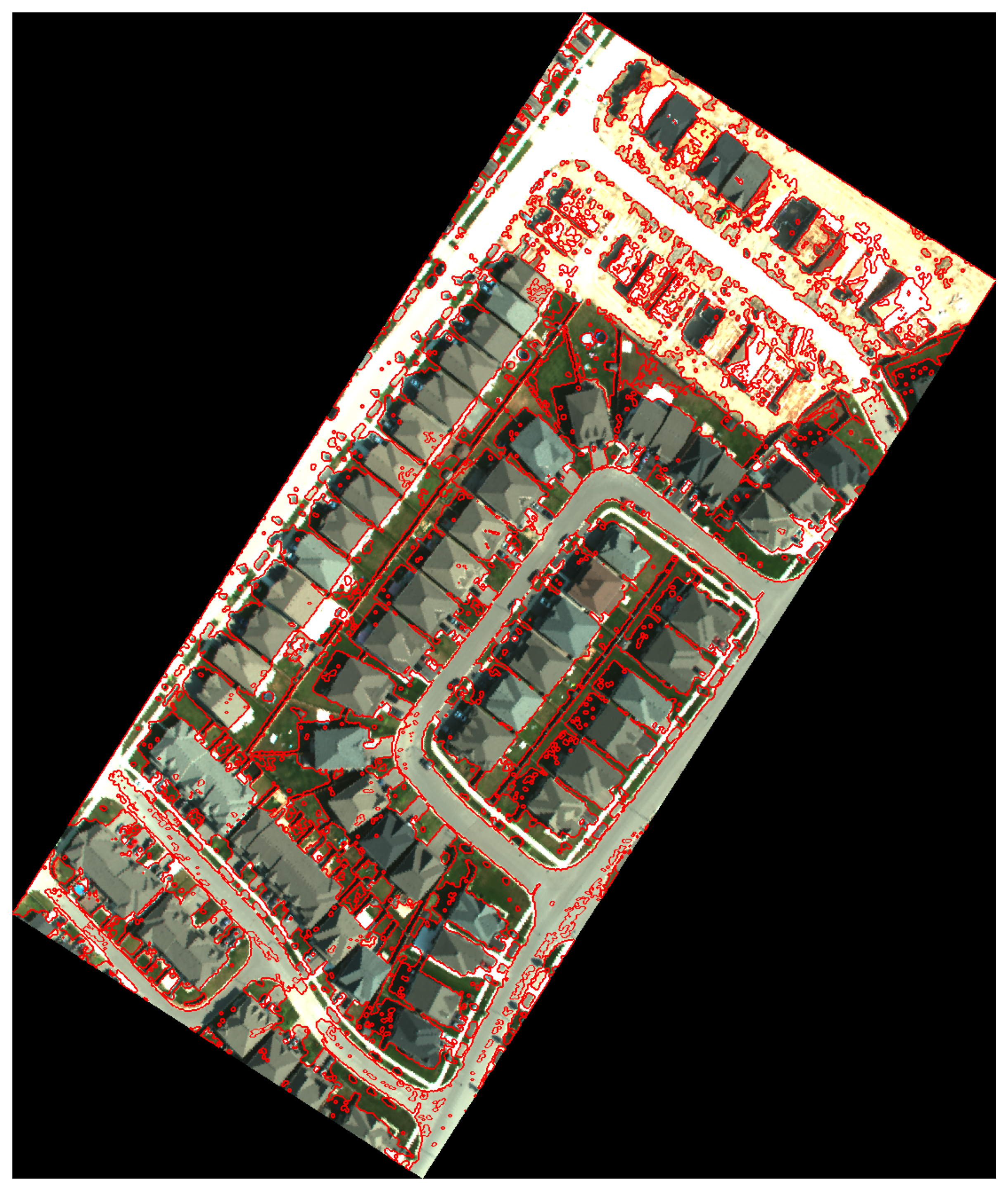}
  \includegraphics[width=0.49\linewidth]{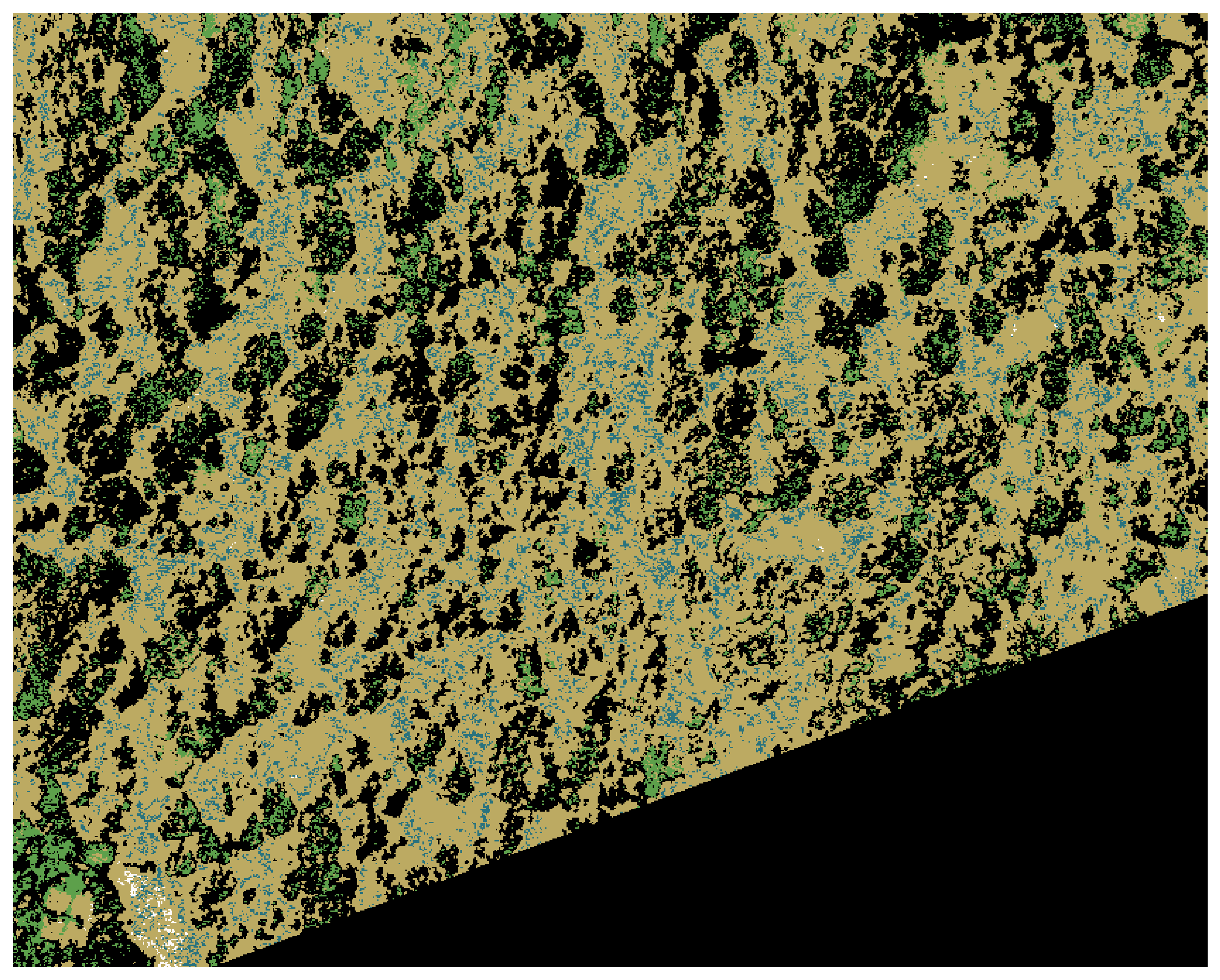}
  \includegraphics[width=0.49\linewidth]{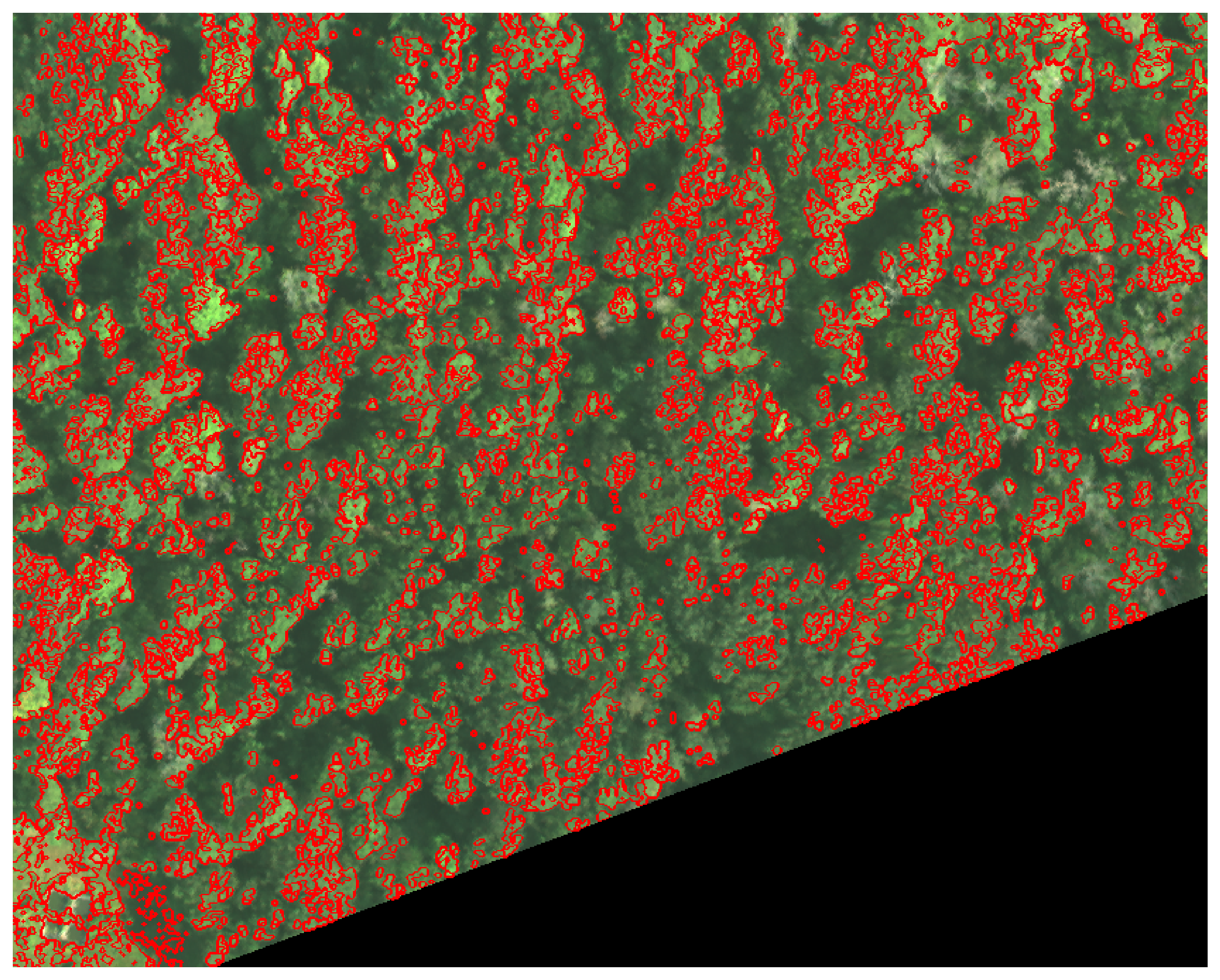}
  \caption{segmentation results using DPGMM algorithm. Left column: objects; right column: boundaries of the segments. From top to bottom: Suburban, Urban, and Forest HSI datasets.}
  \label{fig:sementation_results}
\end{figure}

Figure \ref{fig:sementation_results}  shows, side-by-side, the segments found by
DPGMM (right) and the boundaries of the segments overlaid
on the RGB image (left). We can observe from Figure \ref{fig:sementation_results}  that
the Forest dataset presented the most challenge in classification
due to a large number of shadows which present small values of
reflectance, therefore less distinctive.

\begin{figure}
  \includegraphics[width=0.49\linewidth]{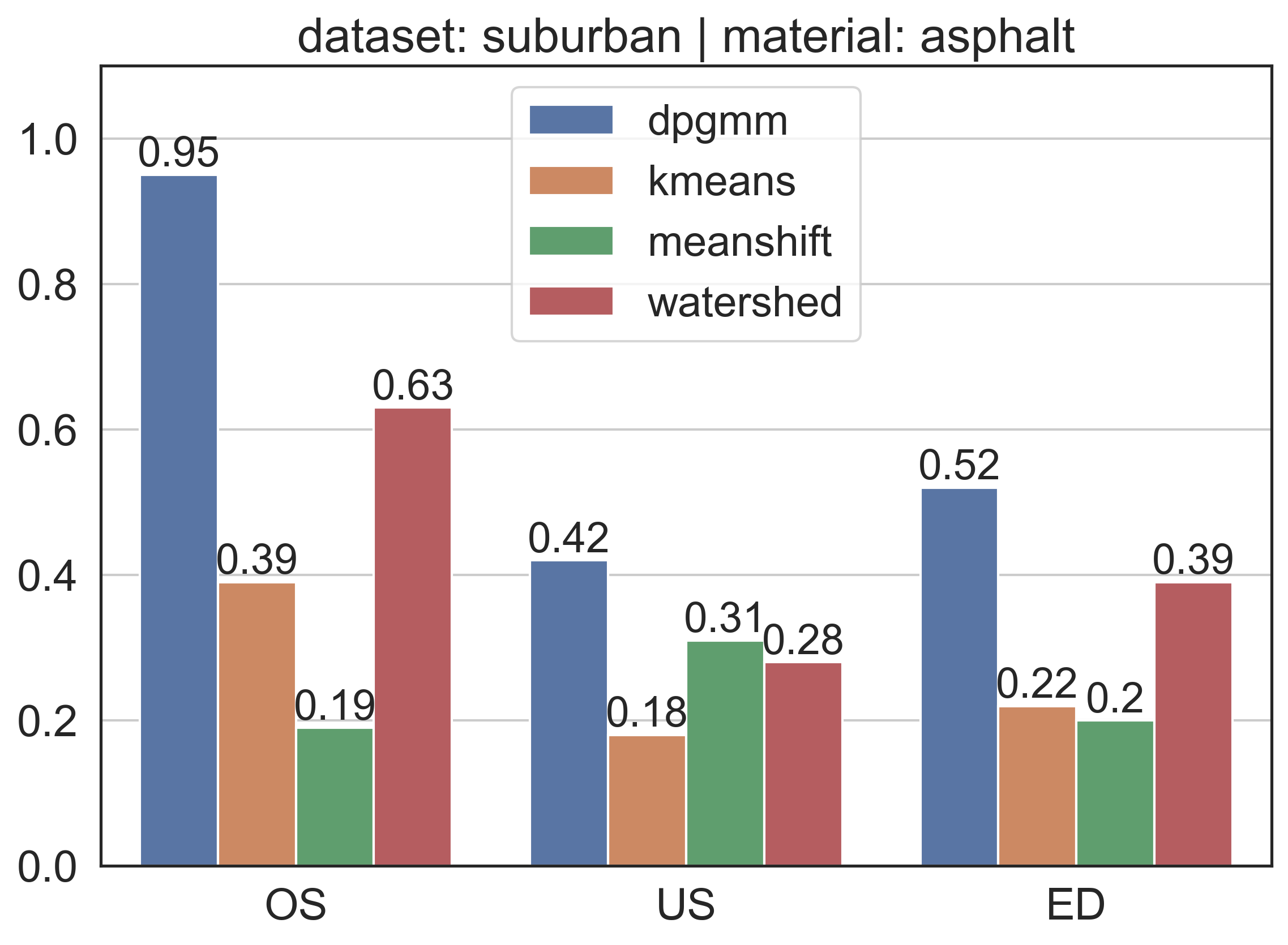}
  \includegraphics[width=0.49\linewidth]{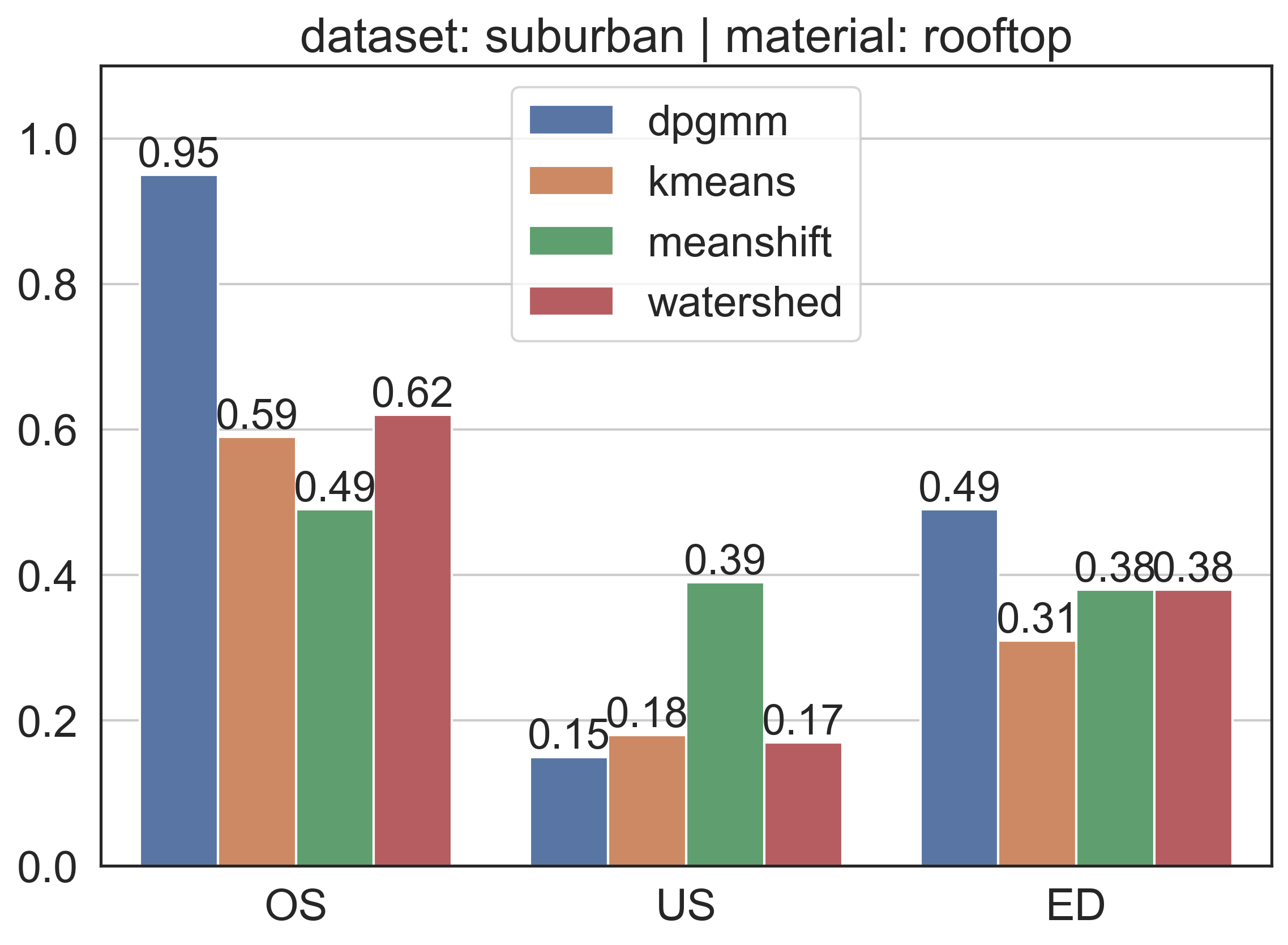}
  \includegraphics[width=0.49\linewidth]{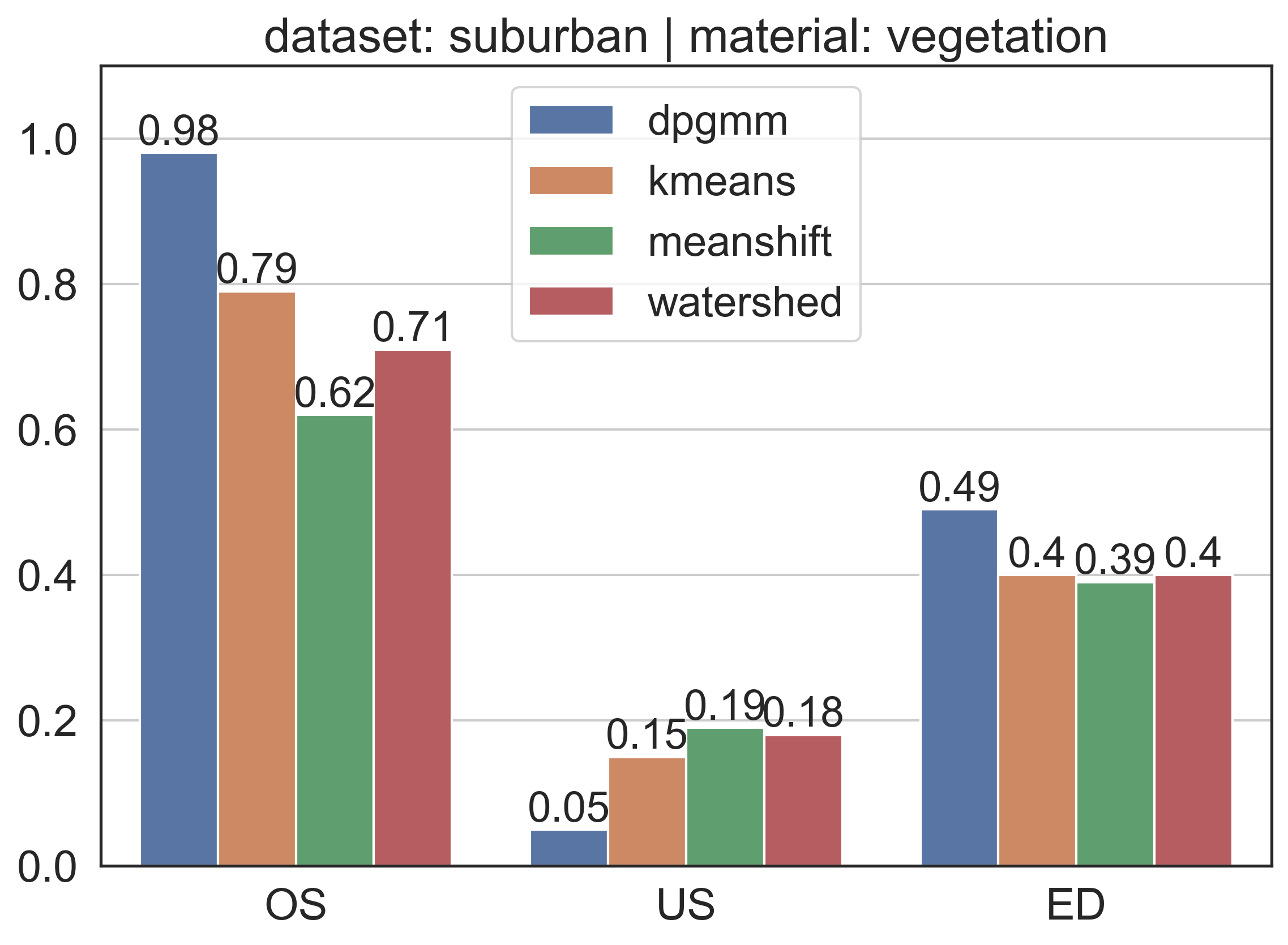}
  \includegraphics[width=0.49\linewidth]{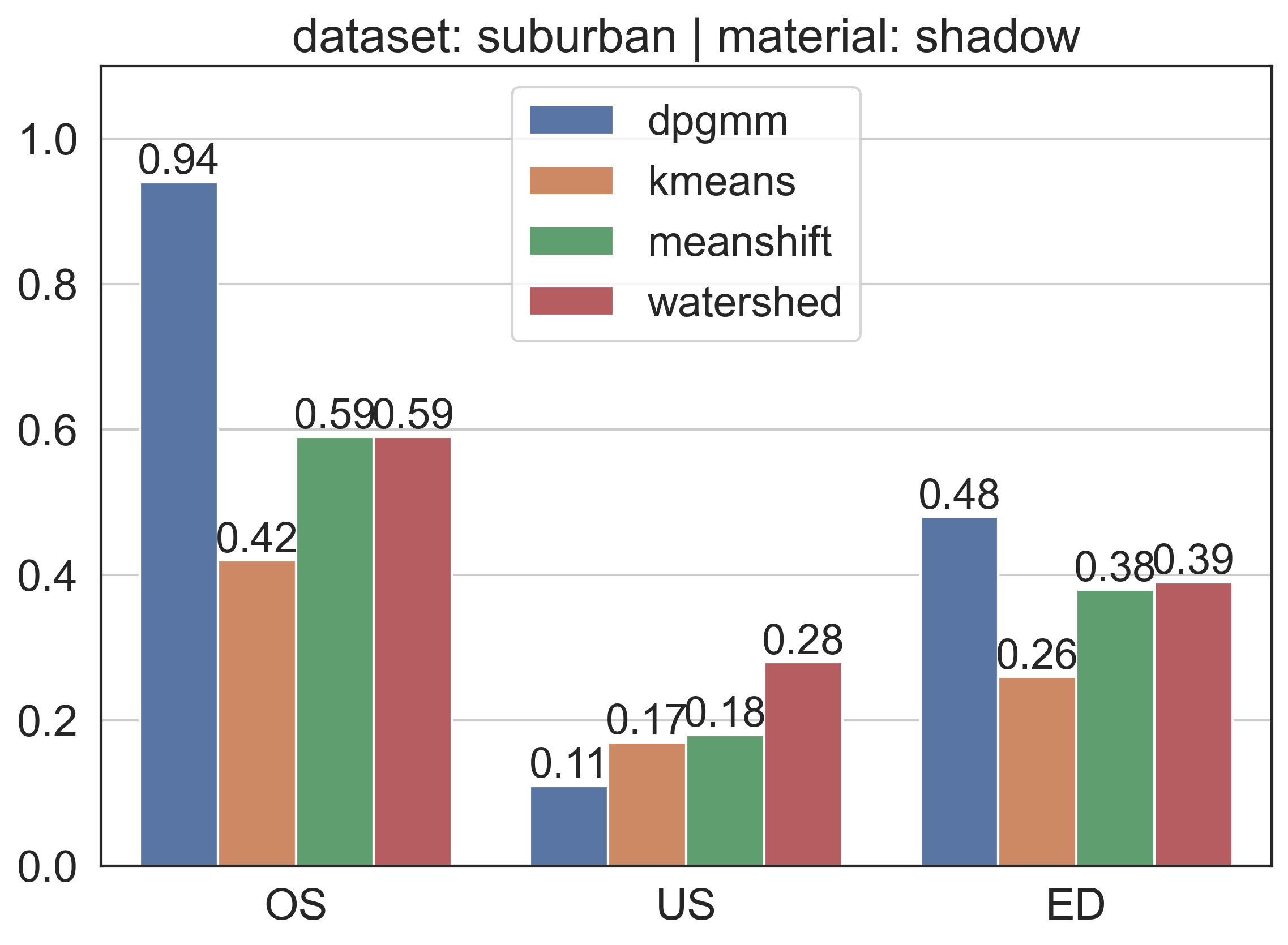}
  \caption{}
  \label{fig:segmentation_metrics_suburban}
\end{figure}

\begin{figure}
  \includegraphics[width=0.49\linewidth]{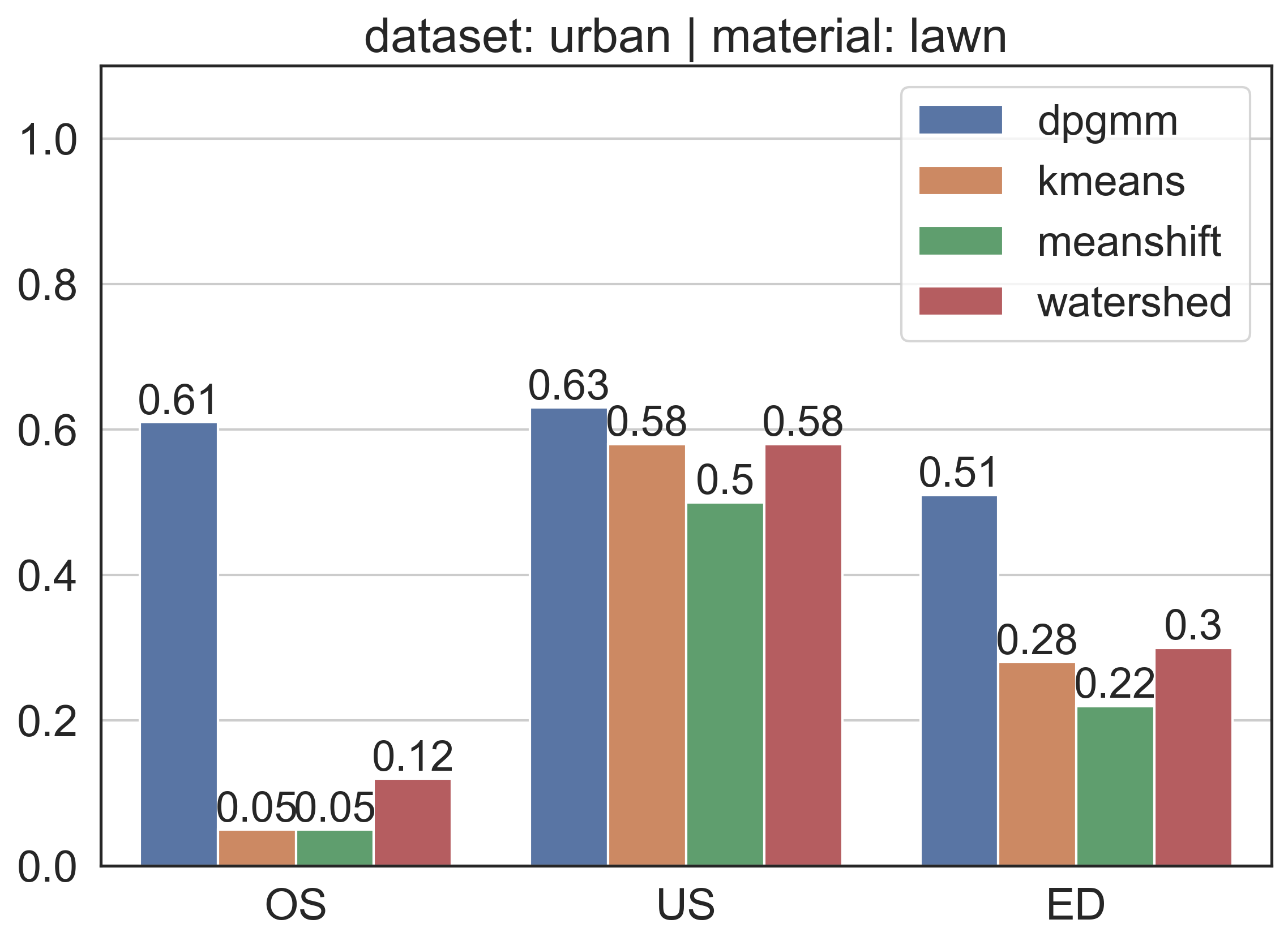}
  \includegraphics[width=0.49\linewidth]{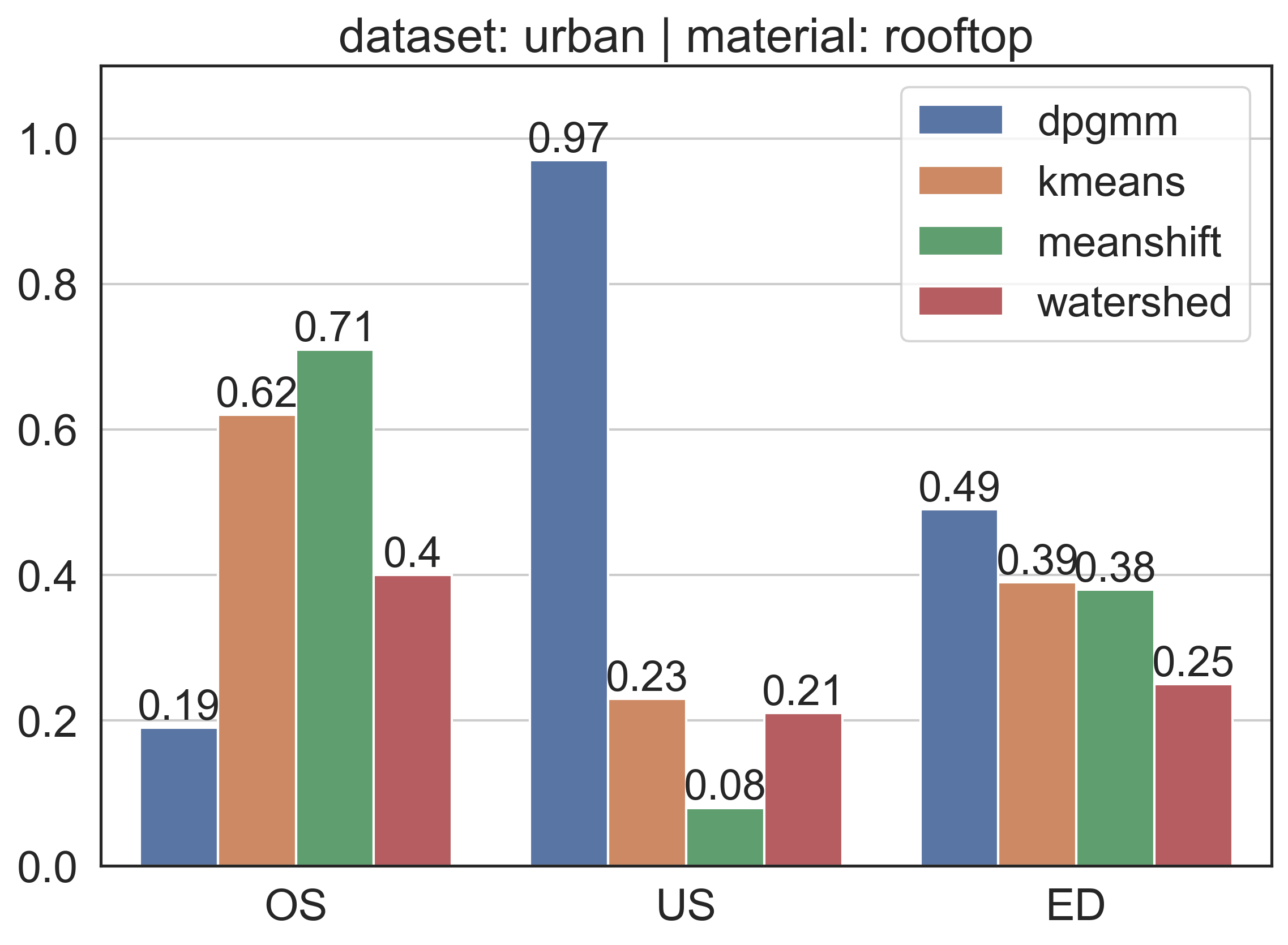}
  \centering{\includegraphics[width=0.49\linewidth]{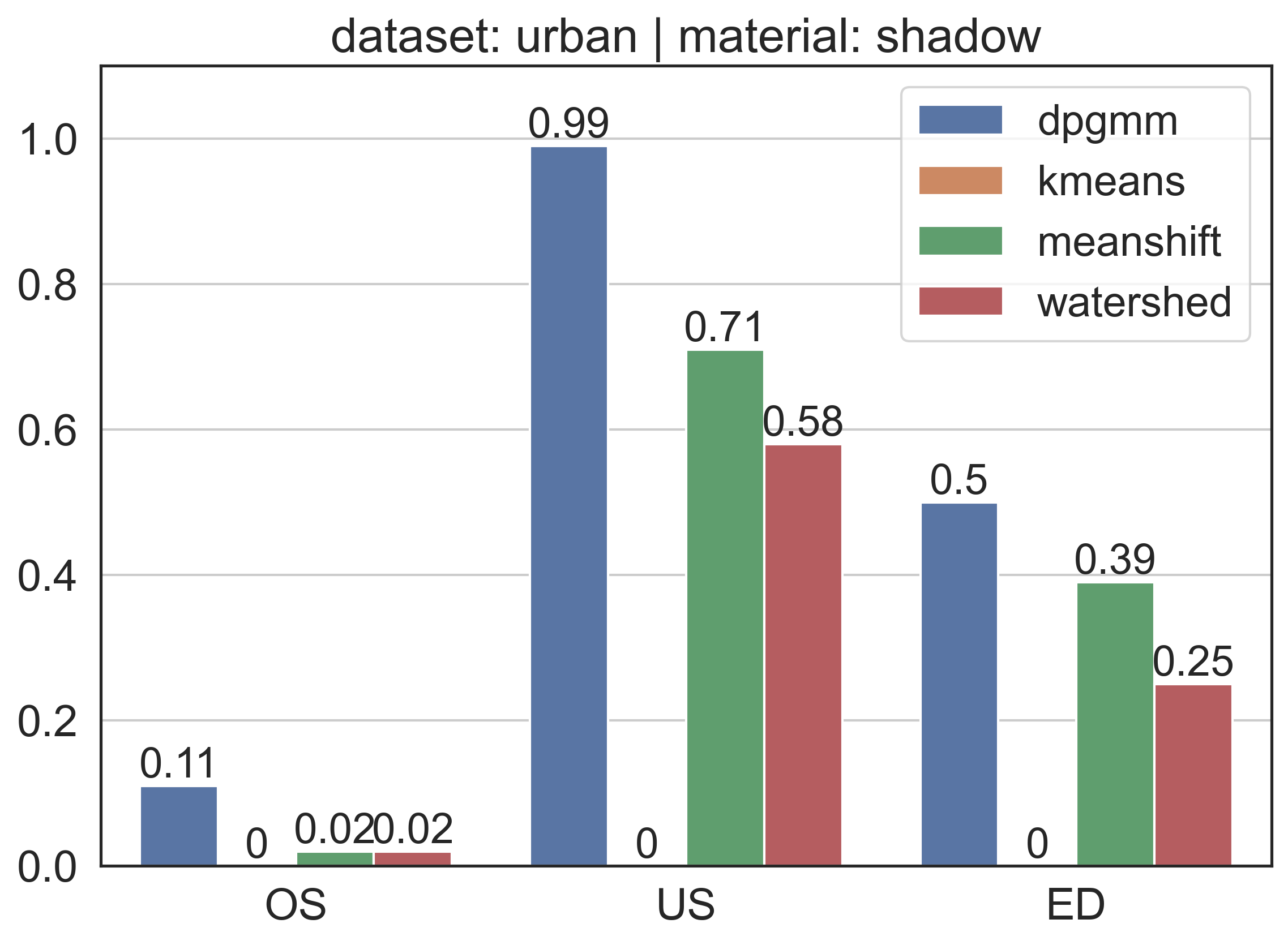}}
  \caption{}
  \label{fig:segmentation_metrics_urban}
\end{figure}

\begin{figure}
  \includegraphics[width=0.49\linewidth]{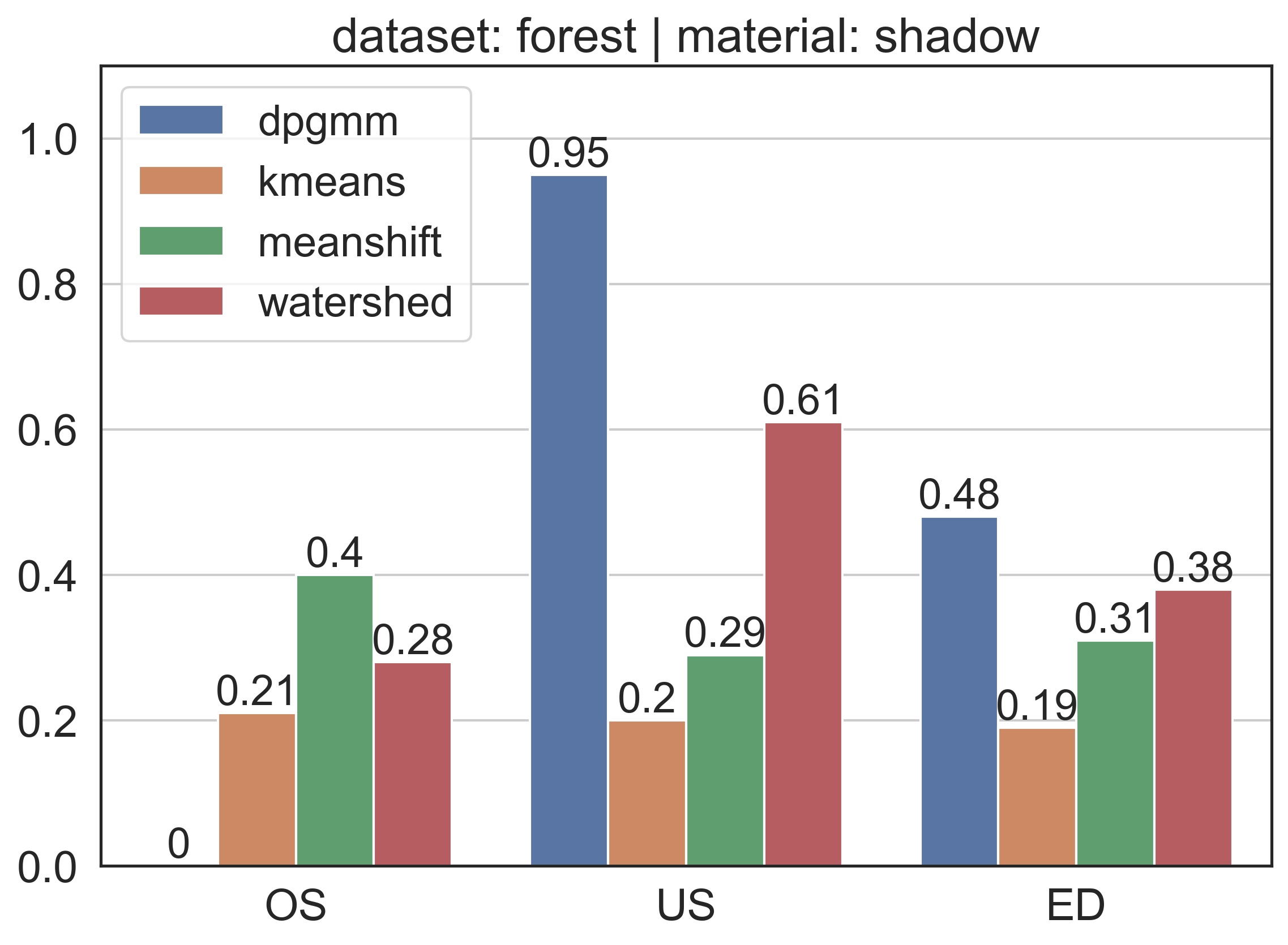}
  \includegraphics[width=0.49\linewidth]{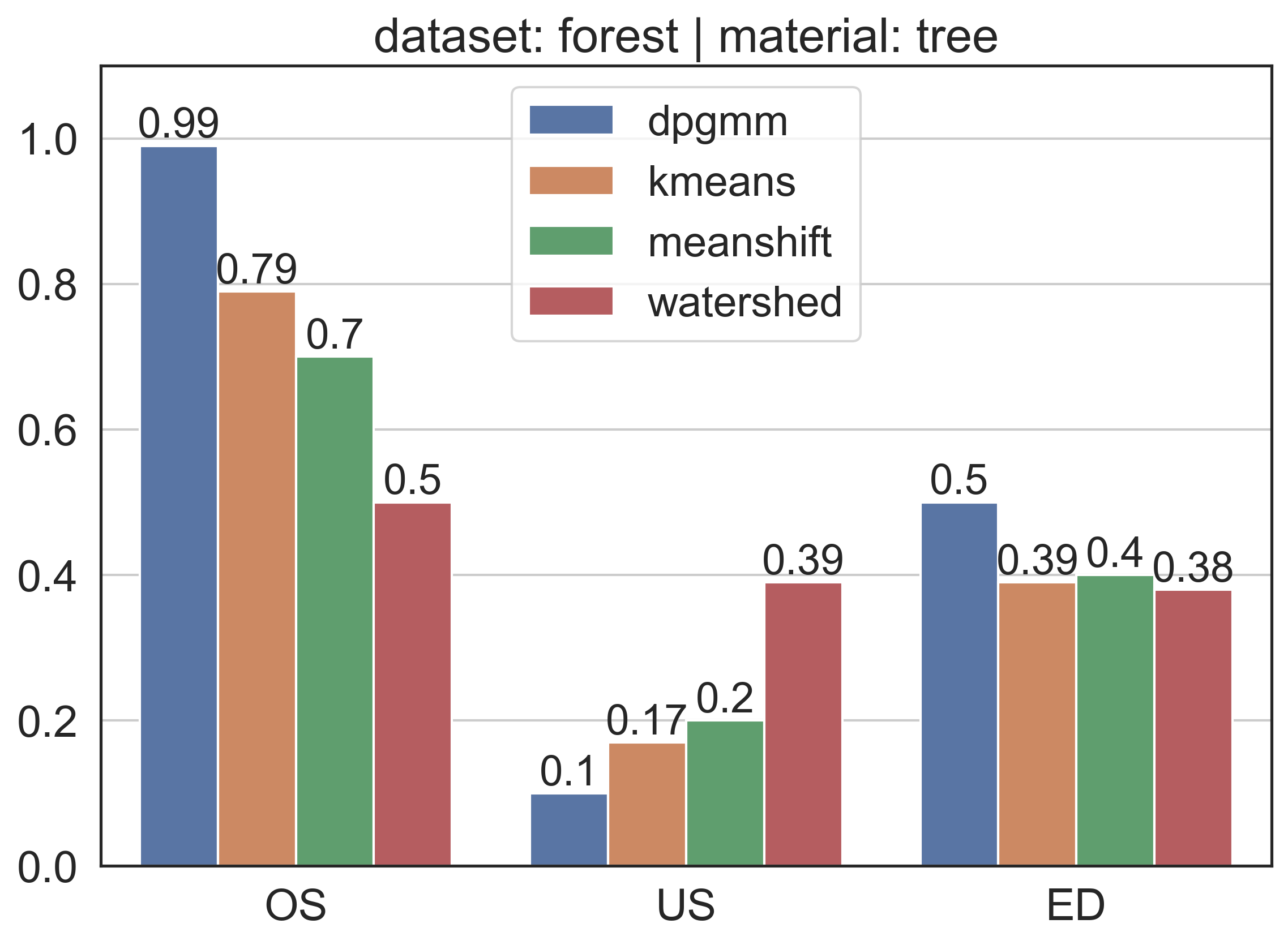}
  \caption{}
  \label{fig:segmentation_metrics_forest}
\end{figure}

Despite excelling in the segmentation metrics and eventually winning state-of-the-art at some instances, DPGMM is not the winner method for all datasets and classes investigated in this
research. However, the overall and qualitative results (Figure ~\ref{fig:average_pixels}) demonstrate that
DPGMM can capture the structure of the pixel spectra while also providing extra information about the mixed
materials in the pixel spectra, and also showcases that this method can be successfully applied in a
multinomial classification task as the next step in a pixel-unmixing pipeline.

\subsection{Execution time}
We also compared the execution time of the DPGMM algorithm to k-means, mean-shift, and watershed. We did not
include the values of the MRS (Multi-resolution segmentation) algorithm due to the considerably larger
execution time, as demonstrated by \cite{dao:2021}. The runtime for inference on DPGMM is two orders of
magnitude faster than the other algorithms, as observed in Figure \ref{fig:execution_time} and Table
\ref{table:execution_time}. The training time is not considered because it is a one-time-only task and does not affect the prediction time. Once the model is trained, the model can be reused several times for inference.


\begin{table}
  \begin{minipage}{0.55\linewidth}
    \includegraphics[width=1\textwidth]{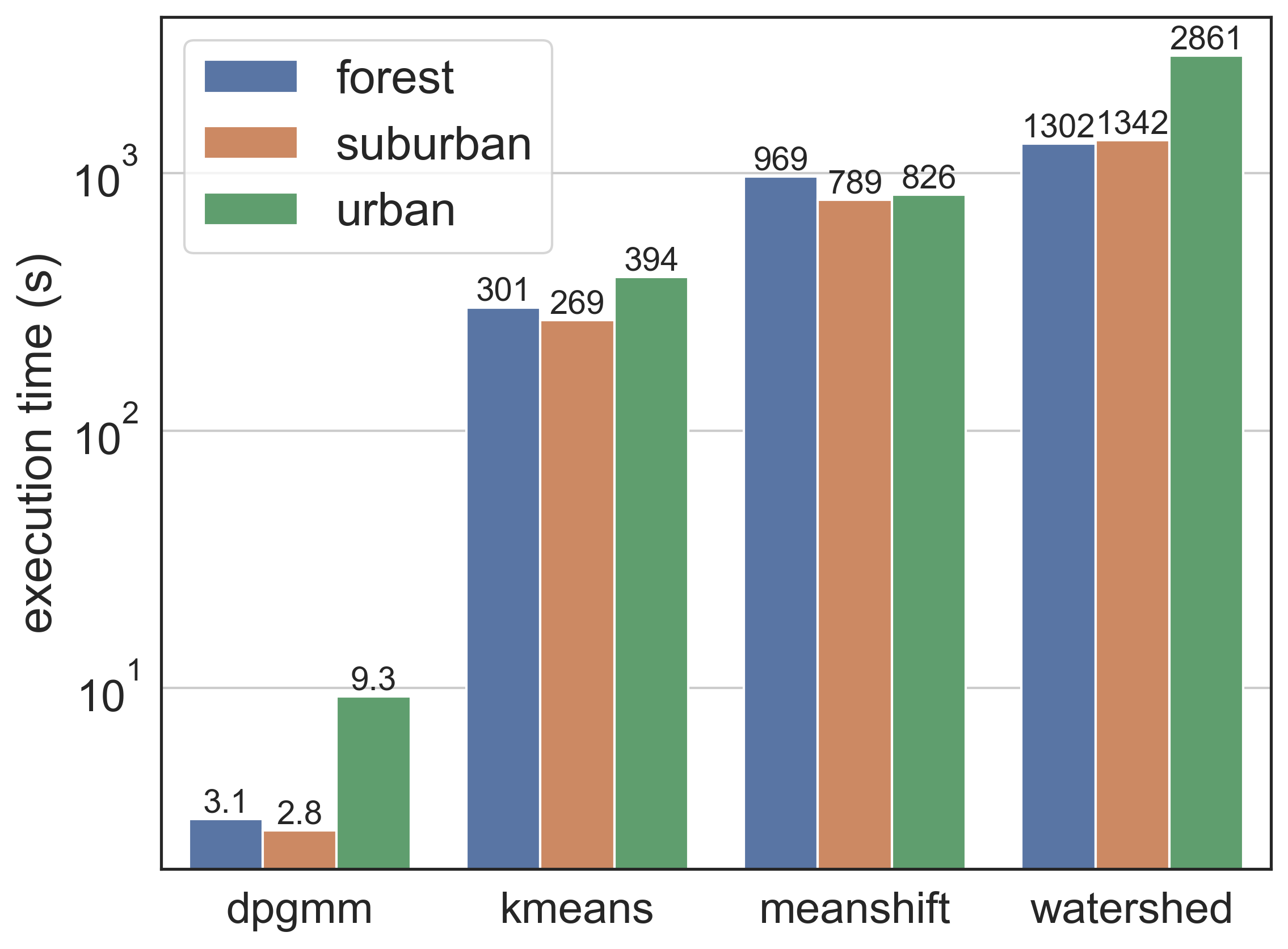}
    \label{fig:execution_time}
  \end{minipage}\hfill
  \begin{minipage}{0.45\linewidth}
    \caption{Execution time (s)}
    \label{table:execution_time}
    \centering
    \scriptsize
    \begin{tabular}{llr}
      \toprule
      {}                           & {}                   & {\textbf{Execution}} \\
      {\textbf{Dataset}}           & {\textbf{Algorithm}} & {\textbf{time (s)}}  \\
      \midrule
      \multirow[c]{4}{*}{Suburban} & dpgmm                & \textbf{2.8}         \\
                                   & kmeans               & 269                  \\
                                   & meanshift            & 789                  \\
                                   & watershed            & 1342                 \\
      \midrule
      \multirow[c]{4}{*}{Urban}    & dpgmm                & \textbf{9.3}         \\
                                   & kmeans               & 394                  \\
                                   & meanshift            & 826                  \\
                                   & watershed            & 2861                 \\
      \midrule
      \multirow[c]{4}{*}{Forest}   & dpgmm                & \textbf{3.1}         \\
                                   & kmeans               & 301                  \\
                                   & meanshift            & 969                  \\
                                   & watershed            & 1302                 \\
      \bottomrule
    \end{tabular}
  \end{minipage}
\end{table}

\section{Conclusion}
\label{conclusion}

We developed a segmentation algorithm based on the Dirichlet Process Gaussian Mixture Model that
automatically finds the scale (or the number of clusters) on spectral features of Hyperspectral Images. We
compared our results with the most common and recent techniques found in the literature of HSI segmentation.
Our results demonstrate
that our method is comparable to the state-of-the-art while also allowing to bypass search for an optimal
scale. While the previous methods require higher runtimes and the evaluation of several parameters and scales, the
algorithm based on Dirichlet Process can find the near-optimal parameters. The qualitative results showed in Figure ~\ref{fig:sementation_results} also indicate that the DPGMM
model is able to capture the structure of the data to identify meaningful segments, which also opens a window
to further extensions of this work in the realm of pixel-unmixing.

\section*{Acknowledgments}
    We acknowledge the support of the Natural Sciences and Engineering Research Council
    of Canada (NSERC) through the NSERC Discovery Program for the funding of the hyperspectral
    image acquisition mission and image preprocessing facility (RGPIN-386183 awarded to Dr. Yuhong He),
    and for the Visual Computing Lab of the Ontario Tech University (RGPIN-2020-05159,
    awarded to Dr. Faisal Z. Qureshi).

\bibliographystyle{unsrtnat}
\bibliography{refs}

\end{document}